\documentclass[conference]{IEEEtran}
\IEEEoverridecommandlockouts
\usepackage{cite}
\usepackage{amsmath,amssymb,amsfonts}
\usepackage{algorithmic}
\usepackage{graphicx}
\usepackage{textcomp}
\usepackage{xcolor}
\usepackage{multirow}
\usepackage{tikz}
\def\BibTeX{{\rm B\kern-.05em{\sc i\kern-.025em b}\kern-.08em
    T\kern-.1667em\lower.7ex\hbox{E}\kern-.125emX}}
\newcommand\copyrighttext{%
  \footnotesize This work has been submitted to the IEEE for possible publication. Copyright may be transferred without notice, after which this version may no longer be accessible.}
\newcommand\copyrightnotice{%
\begin{tikzpicture}[remember picture,overlay]
\node[anchor=south,yshift=10pt] at (current page.south) {\fbox{\parbox{\dimexpr\textwidth-\fboxsep-\fboxrule\relax}{\copyrighttext}}};
\end{tikzpicture}%
}
\begin{document}

\title{On Explaining Unfairness: An Overview}

\author{\IEEEauthorblockN{Christos Fragkathoulas}
\IEEEauthorblockA{\textit{University of Ioannina} \\
\textit{Archimedes / Athena RC} \\
Greece \\
ch.fragkathoulas@athenarc.gr
}
\and
\IEEEauthorblockN{Vasiliki Papanikou}
\IEEEauthorblockA{\textit{University of Ioannina}\\
\textit{Archimedes / Athena RC} \\
Greece \\
v.papanikou@athenarc.gr
}
\and
\IEEEauthorblockN{Danae Pla Karidi}
\IEEEauthorblockA{\textit{Archimedes / Athena RC} \\
Greece \\
danae@athenarc.gr}
\and
\IEEEauthorblockN{Evaggelia Pitoura}
\IEEEauthorblockA{\textit{University of Ioannina}\\
\textit{Archimedes / Athena RC} \\
Greece \\
pitoura@uoi.gr}
}

\maketitle

\begin{abstract}
Algorithmic fairness and explainability are foundational elements for achieving responsible AI. In this paper, we focus on their interplay, a research area that is recently receiving increasing attention. To this end, we first present two comprehensive taxonomies, each representing one of the two complementary fields of study: fairness and explanations. Then, we categorize explanations for fairness into three types: (a) Explanations to enhance fairness metrics, (b) Explanations to help us understand the causes of (un)fairness, and (c) Explanations to assist us in designing methods for mitigating unfairness. Finally, based on our fairness and explanation taxonomies, we present undiscovered literature paths revealing gaps that can serve as valuable insights for future research.
\end{abstract}


\section{Introduction}

Artificial Intelligence (AI) has a profound impact on our everyday lives. As an example, recommendation systems shape to a large extent the decisions we make  about the information and content we consume, the people we engage with, the products we buy, the career paths we follow, and many more. Furthermore, AI algorithms affect decisions at institutional level, such as hiring, sentencing, or policing decisions. For instance, in banking, algorithms play a significant role in determining credit scores, loan approvals, and financial recommendations. In education, they influence student admissions and personalized learning experiences shaping the trajectories of people academic and professional lives.

The broad application of such (semi-) automated decision systems has raised concerns about possible biases that these systems may incorporate. Such concerns are grounded on ample empirical evidence, which demonstrates potentially biased behavior towards minorities or protected groups. For instance, the COMPAS system for recidivism prediction was found to predict higher risk values for black defendants than their actual risk \cite{larson2016we}, models of word embeddings to reflect female/male gender stereotypes \cite{world-embeddings}, health care risk assessment systems to treat black patients unfavorably \cite{racial-health} and job advertisement systems to disproportionately show certain types of job ads to people based on gender \cite{facebook}. Algorithmic fairness aims at mitigating such biases.

Another important aspect of responsible AI is explainability. Explainable AI (XAI) helps end users to understand model decisions, and enables AI developers to scrutinize and rectify model behavior. However, few works have tackled the intersection of explanations and fairness, i.e., explaining (un) fairness. This process differs from traditional explanation approaches since it involves a shift in focus from solely investigating accuracy to investigating the fairness of a model.

Different types of bias can be introduced in the various stages of a machine learning pipeline, such as societal bias in data generation, bias in data selection and collection, and bias in the training of the models. Such biases may be directly related to sensitive attributes, such as gender, race, or ethnicity. Biases may be also due to ``proxy"  attributes that indirectly correlate with sensitive attributes. For instance, features like zip code, personality tests, and resume keywords can act as proxies for race, disabilities, and gender, respectively \cite{ingold2016, kamiran2009classifying, weber2014workplace, dastin2022amazon,le2022survey}. 
Such dependencies highlight the need for explanation techniques that can disentangle these complex relationships.

Effective explanations for fairness should fulfil several key objectives \cite{galhotra2021explaining}: (1) provide mechanisms to evaluate and communicate fairness issues, ensuring different stakeholders that the decision rules of the system are justifiable \cite{selbst2018intuitive}. For instance, to pinpoint cases where individuals are treated differently due to their protected status, even when qualifications or experience are identical or clarify why a specific product recommendation was not made or why a job offer was not extended; (2) offer users an actionable recourse to change the results of algorithms in the future \cite{berk2019machine, karimi2022survey, venkatasubramanian2020philosophical, wachter2017counterfactual}.

In this paper, we begin by examining the multiple notions of fairness and the specific tasks in which it has been applied and then investigate various methodologies in explanations. Fairness and explainability methods are summarized into two comprehensive taxonomies. When it comes to exploring explanations for fairness, we identify three important directions: (a) Explanations to enhance fairness metrics, (b) Explanations to identify and understand the causes of (un)fairness, and (c) Explanations in designing methods for mitigating unfairness. Finally, based on our fairness and explanations taxonomies, we present undiscovered literature paths revealing gaps that can serve as valuable insights for future research.
\copyrightnotice
The remainder of this paper is organized as follows. Section \ref{seq: Fairness} presents core definition of fairness, while Section \ref{sec: Explanations} different explanation approaches. Section \ref{seq: Explanations for Fairness} presents in detail existing work in explanations for fairness. Section \ref{sec: Conclusions and Future Directions} systematically identifies and examines gaps within the existing literature to offer perspectives for future research.

\section{Fairness}
\label{seq: Fairness}
While fairness is a broad term, most formalization of algorithmic fairness focus on non-discrimination, formulated as the
requirement that the results of an algorithm should not be influenced by attributes that are not relevant to the task at hand \cite{pitoura2022fairness, mehrabi2021survey, caton2020fairness}. Such attributes are called protected, or sensitive, and often include among others gender, religion, age, sexual orientation, and race.  

\begin{figure*}[t]
    \centering
    \includegraphics[height=.5\textwidth]{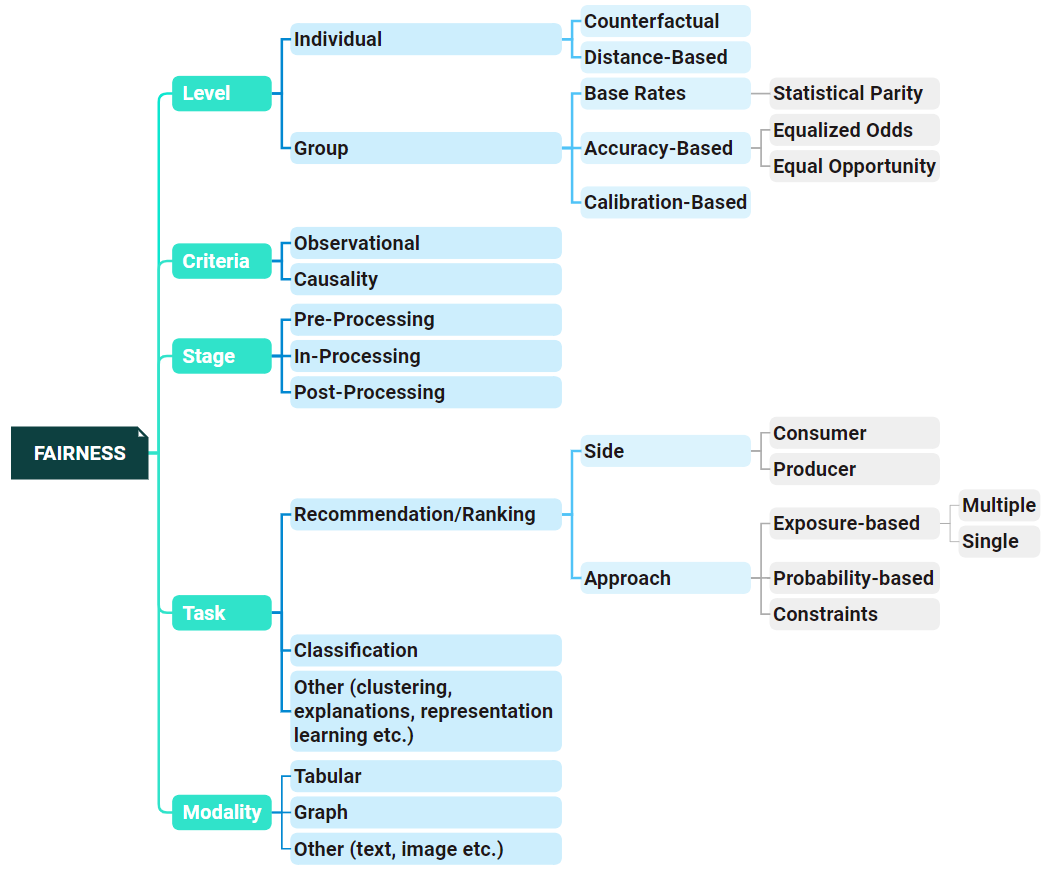}
    \caption{Taxonomy of Fairness Approaches}
    \label{fig: taxonomy on fairness}
\end{figure*}

Since, there has been a proliferation of approaches to modeling, measuring, and addressing fairness, we present a taxonomy to help us place work in explaining unfairness into perspective (see Figure \ref{fig: taxonomy on fairness}). First, when it comes to formalizing unfairness, models of unfairness may be applicable at an individual or group level and may satisfy different criteria. Approaches to mitigating unfairness appear at different stages in the machine learning pipeline. Finally,  models and mitigation approaches depend both on data modality (i.e., tabular, text, image, video, graphs) and  the specific task (i.e., classification, ranking, recommendations, clustering). In this paper, we focus on tabular data and the classification task. We also consider graphs, recommendation and ranking.

\textbf{Level of fairness}
Fairness models are broadly distinguished as individual or group based ones. Individual fairness asks that similar individuals are treated similarly. Group fairness assumes that individuals are partitioned into groups based on the value of one or more of their protected attributes and asks that the groups are treated similarly. We present briefly common formulations of individual and group fairness for classification.

Most individual-based fairness models are distance based where given a distance measure between individual, the difference  of the outputs of the classifier for two individuals is bound by the distance between them \cite{dwork2012fairness}. Alternatively, counterfactual fairness
\cite{kusner2017counterfactual,wu2019counterfactual} posits that an output is fair if it is consistent across both the actual world and a hypothetical scenario where the individual belongs to a different group.

For group fairness, let us assume that there are two groups, the protected group $G^+$  and the non-protected group $G^-$. Let  $f$: $D$ $\rightarrow$ $\{0, 1\}$ be a binary classifier with 1 being the favorable outcome, and $y$, $\hat{y}$ be the ground truth and the predicted output respectively. 
\emph{Base rates} group fairness compares the probability $P(\hat{y}=1 \mid v \in G^+)$ that an individual $v$ receives a favorable outcome when $v$ belongs to the protected group with the corresponding probability $P(\hat{y}=1 \mid v \in G^-)$ when $v$ belongs to the non-protected group. The most commonly used such fairness metric is \textit{statistical parity} that asks that these two probabilities are equal. \emph{Accuracy-based} fairness warrants that various types of classification errors  are equal across groups.
For example, \emph{equal opportunity} expressed as: \(P(\hat{y} = 1|y = 1, v \in G^+) = P(\hat{y} = 1|y=1,  v \in G^-)\) considers true positive rate (TPR), while \textit{equalized odds} requires the same TPR and false positive rate (FPR). Finally, \textit{calibration-based} methods aim at equal calibration across groups in the case of probabilistic classifiers.

\textbf{Fairness criteria}
Fairness definitions can also be approached based on two criteria\cite{castelnovo2022clarification}: \emph{observational criteria}, relying solely on the observational distribution of data, and \emph{causal criteria}, that go beyond observational patterns and seek to understand the underlying causal relationships between variables. This involves incorporating domain knowledge and causal inference techniques to identify and mitigate unfair outcomes caused by underlying causal factors.

\textbf{Stage of fairness}
Methods for mitigating unfairness can be categorized into three distinct groups based on the \emph{stage} at which they are applied \cite{mehrabi2021survey, caton2020fairness}: (a) \emph{Pre-processing} approaches modify the input data,
(b) \emph{In-processing } approaches modify the model so that it produces fair results for any input, and (c) \emph{Post-processing} approaches modify the output.

\textbf{Other tasks and modalities}
In \emph{recommendation and ranking}, 
fairness requirements arise from both the consumer side (users who receive rankings or recommendations) and the producer side (items that are ranked or recommended) \cite{pitoura2022fairness}. In recommendation tasks, we can further separate definitions of fairness into, \emph{exposure-based} fairness, that assesses the expected attention received by a candidate or group of candidates by measuring their average position bias and \emph{probability-based} fairness, that uses statistical methods to determine the probability that a given ranking is the result of a random process \cite{zehlike2021fairness}.
 
\textit{Graph data} with their complex inherent nature (e.g. topological features) introduce unique challenges when addressing fairness, requiring the development of new fairness notions \cite{DongMWCL23}. For example, a topologically biased structure can lead to unfair information propagation between nodes in different subgroups (communities), resulting in biased models for a variety of tasks: representation learning \cite{lahoti2019operationalizing, fan2021fair,oneto2020learning,khajehnejad2022crosswalk, ma2022learning, YDong_GNN, Dai2020SayNT, agarwal2021towards}, classification \cite{jiang2022fmp}, link prediction \cite{10.1007/978-3-030-86520-7_22}. Moreover, in recommendation over biased networks, popular items tend to receive higher scores. This can be mitigated by introducing various diversity-based factors \cite{wasilewski2016incorporating,chen2020esam,abdollahpouri2017controlling,ge2021towards}. Finally, in graph clustering tasks, existing works \cite{gupta2021protecting,kleindessner2019guarantees} focus on ensuring each cluster proportionally represents each sensitive subgroup.

\textbf{Fairness in explanations}
As a final note, besides using explanations for fairness (i.e., the focus of this work), another way that two concepts are connected is by asking for explanations to be fair.
 \emph{Fair explanations} \cite{dai2022fairness, balagopalan2022road, zhao2023fairness} refer to 
 explanations that are accurate and transparent but also align with fairness criteria. In this direction some approaches like \cite{dai2022fairness} compare various metrics that assess the quality of explanations across groups. Such metrics include fidelity (the accuracy in representing the underlying decision-making process), stability (ensuring similar points yield similar explanations), and sparsity (the clarity and ease of interpretation for users). For example, they compare the mean values of these metrics between protected and non-protected groups, with a significant variance indicating disparity.

Another metric of quality for explanations is diversity, commonly applied to recommendations.
For instance, \cite{fu2020fairness} proposed a fairness-aware ranking framework based on knowledge graphs that can be applied on top of existing explainable recommendation algorithms. The method aims to produce explainable recommendations under fairness constraints, focusing on fairness-aware path reranking. 

\section{Explanations} 
\label{sec: Explanations}
\begin{figure*}[t]
        \centering
        \includegraphics[height=.5\textwidth]{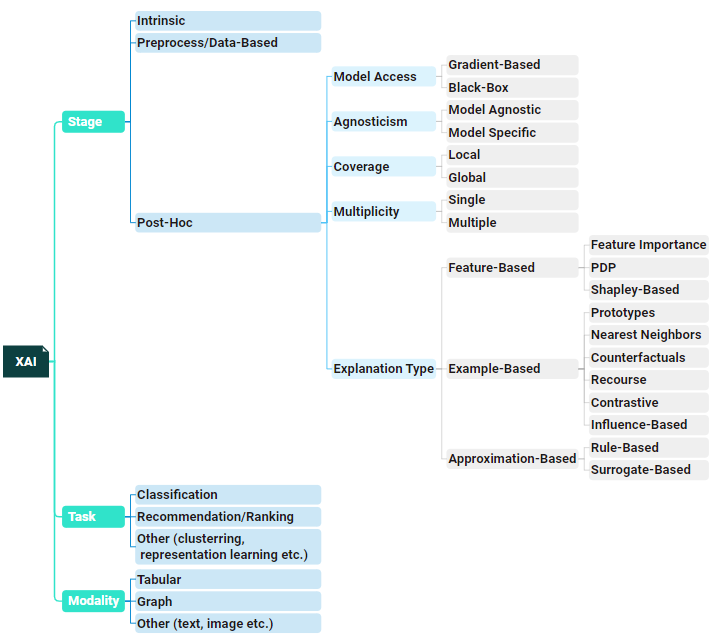}
        \caption{Taxonomy of Explanation Approaches}
        \label{fig: taxonomy on explanation approaches}
    \end{figure*}

Since interpretability and understanding model decisions are vital in machine learning, explainable AI (XAI) has received a lot of attention. To assist us in designing explanations for fairness, in Figure \ref{fig: taxonomy on explanation approaches},  we present a taxonomy that provides a general overview of this broad research field \cite{molnar2020interpretable, verma2020counterfactual, bodria2023benchmarking, dwivedi2023explainable, arrieta2020explainable, adadi2018peeking, guidotti2022counterfactual}. We first classify explanation methods based on the stage in the machine learning pipeline at which they occur as: (a) intrinsic, (b) pre-process or data-based and (c) post-hoc ones. We pay special attention to post-hoc approaches which are the ones most commonly used in explaining unfairness.
    
    \textbf{Intrinsic Methods} are crafted purposefully for transparent and interpretable outcomes \cite{bodria2023benchmarking, adadi2018peeking}. These models are known for their simplicity and inherent transparency, allowing users to understand the reasoning behind each output. However, they may sacrifice accuracy when dealing with complex data patterns \cite{adadi2018peeking, dwivedi2023explainable}.

    \textbf{Preprocess or Data-Based Approaches} 
    utilize unsupervised techniques to uncover hidden data patterns and structures, aiding in enhancing the understanding of the dataset prior to model training. Techniques such as principal component analysis (PCA) \cite{pearson1901liii}, singular value decomposition (SVD) \cite{eckart1936approximation}, non-negative matrix factorization (NMF) \cite{lee2000algorithms}, independent component analysis (ICA) \cite{comon1994independent}, factor analysis (FA) \cite{bartholomew1995spearman}, and t-distributed stochastic neighbor embedding (t-SNE) \cite{van2008visualizing} are commonly employed for dimensionality reduction, feature extraction, and clustering. Analyzing feature correlations provides insights into variable interactions and data structure.

    \textbf{Post-Hoc Approaches}
     are applied after model training to provide explanations for complex machine learning models lacking inherent interpretability. This broad category can be further subdivided based on different criteria.

    \emph{Model Access}
    One subdivision is based on the level of access to the model internals, which significantly impacts how explanations are generated, broadly categorized into:

    \begin{itemize}
        \item \textit{Complete Model Access/White-box:} These approaches offer the algorithm unfettered access to the intricate internal mechanics of the machine learning model \cite{verma2020counterfactual}. This granularity allows for the creation of precise and context-aware explanations, optimizing the potential for achieving high fidelity to the internal operations of the model.         
         
        \item \textit{Access to Gradients:} Explanation methods that utilize gradients, often through optimizers like Adam \cite{kingma2014adam}, are effective when gradient information is accessible. This approach offers a balance between model specificity and versatility, enabling informative explanations without full model access.
        
        \item \textit{Black-box models:} In scenarios where the algorithm operates without direct access to the internal details of the model, it falls under the category of a "black-box" approach \cite{bodria2023benchmarking, verma2020counterfactual, arrieta2020explainable}. Black-box models are generally designed for versatility and adaptability to a wide range of machine learning models, making them suitable for complex, proprietary, or opaque models.
    \end{itemize}

    \textit{Model agnosticism} characterizes the adaptability of an explanation method across different types of machine learning models \cite{bodria2023benchmarking, adadi2018peeking, guidotti2018survey, verma2020counterfactual, dwivedi2023explainable, arrieta2020explainable, guidotti2022counterfactual}. It refers to the capability of the algorithm to operate on a wide range of model architectures without being bound by specific model requirements. 
    While most black box approaches are model agnostic, black box model that requires specific input features or domain knowledge, are not. Conversely, \textit{Model-Specific} \label{seq: Model-Specific} approaches can be used to interpret only a specific type of black-box model, constraining their applicability and versatility to a narrower spectrum of machine learning models. Most of the model specific approaches have gradient or white-box access to the model.

    \emph{Coverage} leads to distinguishing explanations in two key types: global and local explanations \cite{guidotti2022counterfactual, adadi2018peeking, dwivedi2023explainable, verma2020counterfactual, bodria2023benchmarking}. \textit{Global explanations} aim to elucidate the overall logic and decision-making process of a machine learning model. They offer a comprehensive view of how various features influence the predictions of the model on a general scale and are instrumental in validating the alignment of the model with broader expectations and values, such as fairness and bias. \textit{Local explanations} focus on individual predictions, offering specific insights into the reasons behind a particular decision. These are crucial when the objective is to understand and trust the decision of the model in a specific context or to challenge and potentially rectify an individual decision.

    \emph{Multiplicity} refers to whether an approach produces a single or multiple explanations. Single explanations are concise and easy to understand but may not capture the full complexity of the model. Multiple explanations provide a more detailed understanding but can be harder to interpret \cite{verma2020counterfactual}.

    \emph{Explanation Type} refers to the nature of the explanations provided and can be grouped into three main categories: feature-based, example-based, and approximation-based. Approximation and example-based approaches offer different strategies for providing insight into model behavior and decision-making processes for individual predictions \cite{verma2020counterfactual}, while feature-based approaches mainly provide global explanations \cite{dwivedi2023explainable}. 
    
    \textit{Feature-based} explanations \cite{dwivedi2023explainable,arrieta2020explainable} assess the significance of individual features in influencing the prediction of a model. They describe how input features contribute to model output. Feature-based methods, including feature importance \cite{fisher2019all}, partial dependence plots (PDPs) \cite{friedman2001greedy}, and Shapley additive explanations (SHAP) \cite{lundberg2017unified}, assign scores to input features, indicating their relative importance when a prediction is made. 
    
    \textit{Example-based} approaches \cite{verma2020counterfactual, dwivedi2023explainable, adadi2018peeking, guidotti2022counterfactual}, focus on providing explanations through specific instances in the vicinity of the explainee datapoint \cite{verma2020counterfactual}. They encompass a range of techniques such as \textit{counterfactual explanations} \cite{wachter2017counterfactual} determining what should be the change in features of an individual to come across the decision boundary of a classifier and \textit{recourse based} \cite{venkatasubramanian2020philosophical} explanations providing actionable changes for the same purpose \cite{karimi2022survey}. Another approach is using \textit{prototypes} that utilize a set of representative  instances from a class \cite{dwivedi2023explainable}, \textit{nearest neighbors} that leverage neighboring data points \cite{arrieta2020explainable}, and \textit{influence based} methods which trace predictions back to specific training instances that significantly influenced a decision \cite{salimi2022interpretable, brophy2020exit}. Additionally, \textit{contrastive explanations} \cite{karimi2022survey, karimi2021algorithmic}, contrasting the presence and absence of specific features \cite{dwivedi2023explainable}.
    
    Finally, \textit{approximation-based} methods \cite{verma2020counterfactual} aim to approximate the behavior of the underlying model and the decision boundaries in the vicinity of the exlainee datapoint, using interpretable models or simpler proxies. Approaches within this category include \textit{surrogate based} \cite{molnar2020interpretable, guidotti2019factual} explanations, which fit linear models near the explained datapoint, and \textit{rule based} ones \cite{guidotti2019factual}, which extract rules from these models \cite{verma2020counterfactual}.

    \textbf{Task specific explanations}
    Explainability techniques also depend on (a) the data modality (i.e., tabular, text, image, video, graphs, KGs) and (b) the specific task they apply (i.e., classification, ranking, recommendations, clustering). In \textit{classification}, explanations uncover the factors influencing predictions \cite{verma2020counterfactual, adadi2018peeking}. 
    In \textit{recommendation}, explainability refers to providing interpretable reasons for the recommendations generated. Explanations in recommendation employ a variety of methods, including nearest-neighbor, matrix factorization, topic modelling, graph models, deep learning, knowledge reasoning, and association rule mining, as outlined in \cite{zhang2020explainable, tintarev2015explaining}. 
    Explanations also play a crucial role in \textit{graph-based} machine learning, shedding light on the reasoning behind graph operations and outcomes, such as understanding the influential factors in link prediction, node classification, and graph classification tasks \cite{bodria2023benchmarking, prado2022survey}, or prioritizing nodes in recommendations or node classifications. Explanations are integral to graph neural networks (GNNs) \cite{yuan2022explainability} and knowledge graphs, where paths leading to answers serve as explanations for various queries and predictions.

\section{Explanations for Fairness}
\label{seq: Explanations for Fairness}
In this section, we explore how we can use explanations in the context of fairness. 
We identify the following three directions:

\begin{itemize}
\item \textbf{Explanations to enhance fairness metrics:} These methods focus on proposing new metrics to quantify (un)fairness, often by measuring the requirements each individual or group of individuals needs to fulfil to achieve a fair outcome. 
\item \textbf{Explanation methods that help us understand the causes of (un)fairness:} These methods aim to identify the underlying causes contributing to biased results in a model.
\item \textbf{Explanations for designing methods for mitigating unfairness:} These approaches recommend specific actions to mitigate unfairness or propose methods to achieve the enhanced fairness metrics.

\end{itemize}
    
 Most of the current research efforts utilize counterfactual explanations, while a couple of them are feature-based and use Shapley values. Keep in mind that counterfactual explanations are not directly related to the type of fairness called counterfactual fairness \cite{kusner2017counterfactual}. A key distinction is that counterfactual explanations focus on revealing the changes needed for different classification outcomes, while counterfactual fairness aims to assess fairness by considering predictions under different demographic scenarios using causal models \cite{goethals2023precof}.

\subsection{Explaining fairness using counterfactuals}

Given a prediction model $f$ and a data point $\mathbf{x}$, a \textit{counterfactual explanation} is an alternative data point, denoted as $\mathbf{x'}$ that clarifies model predictions by crafting scenarios where the outcome $f(\mathbf{x'})$ differs from the initial prediction $f(\mathbf{x})$ while retaining similarity to the original input. This process involves minimizing the distance between the original and counterfactual inputs through an optimization problem, formulated as:

\begin{equation*}
\mathbf{x'} = {\mathrm{arg}}\ {\mathrm{min}}\ distance(\mathbf{x}, \mathbf{x'}) \quad \text{s.t.} \quad f(\mathbf{x'}) \neq f(\mathbf{x})
\end{equation*}

The basic idea in employing counterfactual for explaining fairness for an individual $\mathbf{x}$  mapped to the negative class ($f(x) = 0$), is to generate its counterfactual $\mathbf{x'}$ so that $\mathbf{x'}$ is mapped to the favorable class ($f(\mathbf{x'}) = 1)$.
This idea is generalized for groups by generating counterfactuals for the set of instances in each group that the classifier mapped to the negative class. Depending on the definition of unfairness, the set of instances may include all instances mapped to the negative class (parity fairness), or the instances falsely mapped to the negative class (error-based fairness). Such counterfactuals can be used to (1) measure unfairness or/and (2) explain the unfavorable behavior of the classifier towards this group.  

%
Let us first discuss how counterfactuals have been used for providing more elaborate definitions of fairness that incorporate the cost of achieving it. 
This cost is often referred as \textit{burden} and is calculated as the distance between the individual and its counterfactual \cite{sharma2020certifai}. Burden for a group $G$ is calculated by averaging the distance between the original input features $\mathbf{x_i}$ and the counterfactual features ${\mathbf{x_i}'}$ for all members of the group under consideration $G$, reflecting the extent of change needed to alter the decision of the model \cite{sharma2020certifai}:

\begin{equation*}
    \text{Burden}(G) = \frac{1}{|G|} \sum_{i \in G} \text{distance}(\mathbf{x_i} - {\mathbf{x_i}'}) \
\end{equation*}

Burden has also been appropriately amortized for group fairness \cite{kuratomi2022measuring}. Building upon this foundational concept, \cite{kuratomi2022measuring} further enriches understanding of burden by integrating it with the false negative rate (FNR) to create a more comprehensive fairness metric, the \textit{normalized accuracy weighted burden (NAWB)}. NAWB for a sensitive group $g$ is calculated as:

\begin{equation*}
\text{NAWB}_s = \frac{\sum_{i \in \mathbf{X}^s_{F\!N}} \text{distance}(\mathbf{x_i}, \mathbf{x_i'})}
            {L \cdot \left|\{\mathbf{x} \in \mathbf{X} | G = g, y = 1\} \right|}
\end{equation*}

where
$L$ is the number of features and  $\left|{\mathbf{x} \in \mathbf{X} | G = g, y = 1} \right|$ the count of instances in the sensitive group $g$ with a positive true label $1$.

A different approach is taken by PreCoF (Predictive Counterfactual Fairness) \cite{goethals2023precof} that uses counterfactuals to understand the causes of unfairness by comparing the relative frequency of the attributes whose values have been changed in the counterfactuals generated for each group.
Both implicit and explicit biases is considered. Explicit bias is evident when the generated counterfactual explanations solely comprise changes in the sensitive attribute, indicating direct discrimination against certain groups. In contrast, implicit bias is identified by removing the sensitive attribute during model training, observing how other attributes influence outcomes differently for protected and non-protected groups. This process uncovers systemic biases that may not be directly linked to sensitive attributes but still impact fairness. 

There are also approaches that provide more elaborate counterfactual explanations for understanding unfairness than identifying important features. To this end, they need to provide a more concise explanation than the set of the generated counterfactuals for all instances in the group. 
Counterfactuals for groups of instances are called  \textit{group counterfactuals}. %
Defining and generating counterfactuals for groups of instances introduces novel challenges. The problem was first introduced in \cite{rawal2020beyond}, where it was modelled as a constraint optimization problem. At a high level, group counterfactuals consist of conditions on feature values, called \textit{actions}, that the instances in the group must satisfy to receive a positive outcome. Recent work includes (a) GLOBE-CE that defines a group counterfactual to be a global direction along which a group of instances may travel to alter their predictions \cite{ley2023globe}, (b) counterfactual explanation trees that build decision trees by assigning actions to multiple instances simultaneously \cite{kanamori2022counterfactual}, and (c) FACTS that uses a frequent itemset approach \cite{kavouras2023fairness}.

FACTS \cite{kavouras2023fairness} (Fairness-Aware Counterfactuals for Subgroups) 
differs from traditional approaches by systematically exploring the space of actions \(A\), measuring the recourse cost \(rc(A, x)\) for an action, determining how many individuals achieve recourse through each action (effectiveness) \(\text{eff}(a, G)\) for individuals in a group $G$ through action $a$, and identifying potential recourse bias among protected subgroups within the feature space. The framework ranks these subspaces based on detected bias and provides interpretable summaries for each, offering a comprehensive and transparent evaluation of fairness. They posit that a classifier can be deemed fair if it satisfies certain criteria, like equal effectiveness and equal choice of recourse. The first one means that the same proportion of individuals in the protected group $G^+$ and in the non-protected $G^-$ can achieve recourse, while the second says that both groups have an equal choice of sufficiently effective actions for achieving recourse, with 'sufficiently effective' defined as actions working for at least 100$\phi$\% of the subgroup for \(\phi\) \(\in [0, 1]\):

\begin{align*}
    \text{aeff}(A, G^+) &= \text{aeff}(A, G^-)\\
    |\{ a \in A \, | \, \text{eff}(a, G+) &\geq \phi \}| = |\{ a \in A \, | \, \text{eff}(a, G^-) \geq \phi \}|
\end{align*}

An issue with counterfactuals is that it is unclear whether the suggested actions can be used to design mitigation actions. Specifically, the suggested shifts in feature values may not be feasible. Additional constraints could be considered, for example, avoiding updates of immutable attributes \cite{ustun2019actionable}. However, this may still be insufficient since there may be casual dependencies between attributes that need to be taken into account. 

To address this problem, instead of interpreting actions as independent feature manipulations, \textit{actionable recourse} \cite{karimi2021algorithmic} suggests that actions should be interpreted as interventions, termed \textit{flipsets}, in a casual model of the world. 
Given a structural causal model (SCM), actions are modelled as structural interventions, which can be thought of as transformations between SCMs. 
The following optimization objective is formulated:

\begin{equation*}
    \begin{aligned}
        A^* \in \arg\min_{A \in \mathcal{A}} \text{cost}(A; \mathbf{x})
            & \quad \text{subject to} \quad f(\mathbf{x'}) \neq f(\mathbf{x})\\
            & \quad \mathbf{x'} = F_A(F^{-1}(\mathbf{x}))\\
            & \quad \mathbf{x'} \in \mathcal{P}, \quad A \in \mathcal{F},
    \end{aligned}
\end{equation*}

where $A$ is the set of interventions that can be constructed as $A = \text{do} (\{X_i := a_i \} _{i \in I})$,  the ``do'' operator replaces the structural equation in the SCM when is applied to a variable $X_i$, $A^* \in \mathcal{F}$ is the minimal set of actions for the minimally costly recourse, $\text{cost}(\cdot ; \mathbf{x}_{\text{F}}) : \mathcal{F} \times X \rightarrow \mathbb{R}^+$ is the cost function and $\mathbf{x'} = F_{A^*} (F^{-1} (\mathbf{x}))$ is the resulting structural counterfactual and $P$ refers to plausibility constraints.

In \cite{karimi2022survey},
a distinction is based between \textit{contrastive explanations} that are explanations used to understand unfairness (e.g., what input changes would lead to a favorable outcome) and \textit{consequential recommendations} used to mitigate fairness (e.g., also actions would lead to the specific changes). A counterfactual-based explanation is a nearest contrastive explanation and an actionable recourse a consequential recommendation.

Another approach to incorporating causality is offered by the \textit{probabilistic contrastive counterfactuals} that exploit probabilistic causal models \cite{galhotra2021explaining}. Actions are in the form of intervention queries. 
Unlike actionable recourse, intervention queries can be estimated from historical data, thereby making no assumptions about the structural equations in the underlying probabilistic causal models. 

Finally, a different approach is taken in \cite{gupta2019equalizing} where an individual recourse is defined as the distance of the individual from the decision boundary of a classifier and group recourse as the average recourse of all individuals in the group \cite{gupta2019equalizing}. A classifier is presented that uses a regularized objective to minimize the difference in recourse across groups. An alternative to this distance-based definition of recourse is proposed in \cite{von2022fairness} that takes into account the causal relationships between features \cite{von2022fairness}.
Specifically, following a counterfactual-based fairness definition, they define that a (causal) recourse may be considered fair at the level of the individual if the cost of recourse would have been the same had the individual belonged to a different group. They also study theoretically and empirically how to enforce fair causal recourse by altering the classifier. 

\subsection{Explaining fairness using Shapley values}
The Shapley method is used to explain the contribution of the features to the accuracy of the model and involves retraining a model on various feature subsets, denoted as $S \subseteq F$, where $F$ represents the set of all features. Each feature is assigned an importance value, indicating its impact on the model prediction when included. The method computes this impact by training two models: $f_{S \cup {i}}$ with the feature included and $f_S$ with the feature withheld. Predictions from these models are compared on a specific input, resulting in differences. These differences are computed for all possible subsets $S \subseteq F \setminus {i}$, capturing the interdependence of withheld features. The Shapley values, denoted as $\phi_i$, are then calculated as a weighted average of these differences using the following formula: 

\begin{equation*}
\phi_i = \sum_{S \subseteq F \setminus \{i\}} \frac{|S|!(|F| - |S| -1)!}{|F|!} \left [f_{S \cup \{i\}}(x_{S \cup \{i\}}) - f_S(x_S) \right]
\end{equation*}

For explaining unfairness, Shapley-based approaches leverage the global Shapley value not to explain the contribution of the features to the output accuracy of the model, but instead for their contribution to the parity fairness of the model \cite{begley2020explainability}. 
The first step is to replace the model  $f_{S \cup {i}}$  and $f_S$ with a value function that captures the fairness effect that quantifies the impact on the fairness measure of including or excluding specific features \cite{begley2020explainability}.

However, decomposing model disparity into feature contributions ignores the possible causal relationships between features. To tackle this problem, FACTS (FairnessAware Causal paTh decompoSition) proposes a framework that decomposes the model disparity as the sum over the contributions of a set of causal paths (instead of features) linking the sensitive attributes with the outcome variable \cite{pan2021explaining}.

Finally, a different approach is taken by data-based explanations \cite{salimi2022interpretable,zhu2022generating}. Here, the goal is to identify the set of instances that, if removed (or updated) from the training data, will cause the largest reduction in classification unfairness. The instances are described using patterns, which are conjunctions of bounds on the feature values. 

\subsection{Explaining fairness beyond classification}

\noindent\textbf{Recommendations.}
Counterfactual explanations in the context of fairness have been expanded within recommendation systems. In \cite{Zafiriou2023}, authors utilize the RecWalk algorithm \cite{nikolakopoulos2019recwalk} to estimate user-item scores via random walks on a graph, focusing on personalized recommendations and evaluating the impact of edge removals on estimated ratings. They provide explanations at both the user and item levels, including user group and item group scenarios. CFairER \cite{wang2023counterfactual}, addresses fairness disparities in recommendation systems by generating attribute-level counterfactual explanations. It employs off-policy reinforcement learning via a Heterogeneous Information Network (HIN) to identify minimal attribute sets that can improve fairness, combining reinforcement learning with attentive action pruning to offer precise insights into factors influencing fairness in recommendation models. CEF \cite{ge2022explainable} generates fairness explanations by formulating an optimization problem to learn the “minimal” change of the input features that changes the recommendation results to a certain level of fairness. Based on this counterfactual recommendation result of each feature, CEF calculates an explainability score in terms of the fairness-utility trade-off to rank all the feature-based explanations and select the top ones as fairness explanations. In terms of explaining unfairness in ranking, the only related work we are aware of is the very recent work Dexer that uses Shapley values to identify attributes that significantly affect the ranking of a group \cite{moskovitch2023dexer}. Then, CEF visualizes the value distribution of such attributes to analyze the difference between the detected group and top-k ranked tuples.

\noindent\textbf{Graphs.} 
Explaining fairness extends to other data modalities, such as graphs, with the current literature limited to research in the context of GNNs and knowledge graphs, namely: (a) \cite{dong2022structural} that explains unfairness in GNNs by identifying for each node two edge sets in its computational graph that can maximally account for the exhibited bias and maximally contribute to the fairness level of the node prediction, (b) \cite{dong2023interpreting} that proposes a strategy to measure the bias exhibited in GNNs, and develop an algorithm to efficiently estimate the influence of each training node on such bias, (c) GNNUERS \cite{medda2023gnnuers} that explains unfairness in GNN-based recommender systems by perturbing the original bipartite graph of user-item interactions to identify which interactions lead to user unfairness, and (d) the work in \cite{fu2020fairness} that mitigates the bias arising from different user activity levels by re-ranking explainable recommendations (KG paths) based on path distribution between users to satisfy specific fairness constraints and deliver diverse explanations.

\begin{table*}[htbp]
\caption{Overview of Approaches for Explaining (Un)Fairness}
\begin{center}
\begin{tabular}{|c|c|c|c|c|c|c|c|c|c|c|}
\hline
 & \multicolumn{6}{c|}{\textbf{Explanation}} & \multicolumn{2}{c|}{\textbf{Fairness}} & &\\
\cline{2-9} 
\textbf{Appr.} & \textbf{\textit{Stage}} & \textbf{\textit{Access}} & \textbf{\textit{Agnostic}} & \textbf{\textit{Coverage}}  & \textbf{\textit{Type}} & \textbf{\textit{Output}} & \textbf{\textit{Level}} & \textbf{\textit{Type}} & \textbf{\textit{Task}} &  \textbf{Goal} \\ \hline

\sloppy\cite{galhotra2021explaining} & Post & B & A & Both & Contrastive & \multirow{2}{*}{\begin{tabular}{@{}c@{}}Propabilistic contrastive\\ CFEs  actionable recourses\end{tabular}} & Both & Fairness of recourse &  Clf& U \\ 
& & & & & & & & & & \\ \hline

\sloppy\cite{salimi2022interpretable} & Post & G & S & G &  Influence-based & Predicate-based causal & Group & Base-Rates/Accuracy-Based & Clf& U, M \\ \hline
\sloppy\cite{goethals2023precof} & Post & B & A & L & CFE & \multirow{2}{*}{\begin{tabular}{@{}c@{}}Most significant\\ feature change\end{tabular}} & Group & Implicit/Explicit bias & Clf & U\\ 
& & & & & & & & & & \\ \hline
\sloppy\cite{sharma2020certifai} & Post & B & A & L & CFE & CFEs & Both & Burden & Clf& E, U \\ \hline
\sloppy\cite{kuratomi2022measuring} & Post & B & A & G & CFE & Burden & Both & Burden & Clf& E, U \\ \hline
\sloppy\cite{rawal2020beyond} & Post & B & A & Both & Recourse & Two level Recourse Sets & Both & User study & Clf& U \\ \hline
\sloppy\cite{ley2023globe} & Post & B & A & G & CFE & CFEs & Group & Fairness of recourse & Clf & U \\ \hline
\sloppy\cite{kavouras2023fairness} & Post & B & A & G & CFE & CFEs & Group & Fairness of recourse  & Clf& E, U \\ \hline
\sloppy\cite{pan2021explaining} & Post & B & A & G & Recourse & Causal path & Group & Base-Rates  &  Clf& U, M \\ \hline
\sloppy\cite{gupta2019equalizing} & Post  & B & A & G & Recourse & Recourses & Group & Fairness of recourse & Clf &  E, M  \\ \hline 
\sloppy\cite{von2022fairness} & Post & B & A & Both & Recourse & Recourses & Both & Fairness of recourse & Clf & E, M \\ \hline 
\sloppy\cite{dong2022structural} & Post & B & A & L & CFE & Edge-Set & Both & \multirow{2}{*}{\begin{tabular}{@{}c@{}}Dist/on Distance-Based \\ Base-Rates/Accuracy-Based\end{tabular}} & Clf & E, U, M\\ 
& & & & & & & & & & \\ \hline
\sloppy\cite{begley2020explainability} & Post & B & A &  Both & Shapley & \multirow{2}{*}{\begin{tabular}{@{}c@{}}Shapley based \\visualization\end{tabular}} & Group & Base-Rates& Clf& U, M \\
& & & & & & & & & & \\ \hline
\sloppy\cite{Zafiriou2023}  & Post & B & A & Both & CFE & CFEs & Both & Base-Rates & Recs & U \\ \hline
\sloppy\cite{wang2023counterfactual} & Post & B & A & G & CFE & CFEs & Group& Exposure & Recs & U, M \\ \hline
\sloppy\cite{ge2022explainable}  & Post & B & A & G & CFE & CFEs & Group & Exposure & Recs& U, M \\ \hline
\sloppy\cite{moskovitch2023dexer} & Post & B & A & G & Shapley & \multirow{2}{*}{\begin{tabular}{@{}c@{}}Attribute Shapley value\\ distribution visualization\end{tabular}} & Group & Exposure & Rank & U \\ 
& & & & & & & & & & \\ \hline
\sloppy\cite{dong2023interpreting} & Post & G & S & G & Influence-based & Node influence & Group & Base-Rates/Accuracy-Based& Clf& E, U, M \\ \hline
\sloppy\cite{zhu2022generating} & Post & B & A & G & Contrastive & Top-k data subsets & Group & Base-Rates/Accuracy-Based & Clf& U, M \\ \hline
\sloppy\cite{medda2023gnnuers} & Post & B & A & G & CFE & CFE & Group & Exposure & Recs& U, M \\ \hline
\sloppy\cite{fu2020fairness} & Post & B & A & Both & Example-based & Topk-k KG-path & Both & Constraints & Recs& E, U, M \\ \hline
\end{tabular}
\label{tab:approach_features}
\end{center}
\end{table*}

\section{Summary and Future Directions}
\label{sec: Conclusions and Future Directions}
In this paper, we first developed taxonomies for fairness and explanation methods, categorizing existing literature based on key dimensions. We then explored the intersection of fairness and explainability. 
Table \ref{tab:approach_features} presents a comprehensive comparison of the various approaches developed to explain fairness. It can serve as a detailed guide for understanding the landscape of fairness explanation methods, highlighting their diversity in methodology and objectives. 

In terms of \textit{explanations} (see Figure \ref{fig: taxonomy on explanation approaches}), Table \ref{tab:approach_features} categorizes whether approaches utilize inherently explainable models, explain the data before model training, or explain non-transparent models (Post). It specifies their access type, whether model-agnostic (A) or model-specific (S), determining applicability across models. In terms of coverage, it indicates whether explanations offer a holistic view (G) or case-specific insights (L). Then, the explanation type (e.g., CFE, Shapley values) and output format (e.g., structural change, edge set) of each approach are listed, providing insights into how explanations are presented. In terms of  \textit{fairness} (see Figure \ref{fig: taxonomy on fairness}), Table \ref{tab:approach_features} differentiates between individual and group-level fairness based on sensitive attributes. It identifies the fairness criteria each approach aims to satisfy, including statistical group fairness measures and other notions such as fairness of recourses and burden. The table also specifies the \textit{task} and \textit{goal} for each approach, such as classification (Clf) or recommendation systems (Recs) and its ultimate objective (e.g., enhancing fairness metrics (E), providing causal understanding (U), mitigating unfairness (M).

A comprehensive overview of Table \ref{tab:approach_features} reveals that a predominant trend among approaches for explaining fairness is the adoption of post-processing methods, indicating a common emphasis on explaining models after training. Furthermore, most of these methods are black-box, model-agnostic, and geared towards addressing global explanations for the models. Notably, counterfactual explanations (CFEs) emerge as a prevalent technique employed by these approaches, while the primary focus often centres around group fairness.

A first observation is that 
most current research focuses on a specific type of fairness. It would be interesting to explore their generality in terms of their applicability to other forms of fairness as well.
There is also room for exploration into dynamic fairness criteria that adapt to evolving data distributions and societal norms. This could include the development of fairness metrics and explanations that are responsive to the changing landscape of data and demographics.
Furthermore, most approaches tend to focus either on individual or group levels of fairness. Therefore, future directions in this field could involve bridging the gap between individual and group levels, thus offering a more comprehensive view of fairness, akin to the advancements made by  \cite{rawal2020beyond, ley2023globe}. Such mechanisms could implement hybrid models that balance the nuances of individual cases with the broader patterns observed at the group level, ensuring more equitable treatment across the board.

Furthermore, current research is limited in its coverage of various ML tasks. While efforts have been made to explain fairness in binary classification and top-k recommendation scenarios, there is a significant gap in extending such work to more complex tasks like multiclass classification, regression, clustering, representation learning, unsupervised learning and applications involving graphs and knowledge graphs.  

From another perspective, developing methods with the capacity to generate diverse explanations (e.g., multiple explanations of the same type, or/and explanations of more than one type) would empower users with a broader range of resources (set of actions), enhance their ability to comprehend and act upon explanations, and potentially enable more customized and efficient approaches to addressing fairness concerns. 
Another direction is to leverage new metrics that provide insights into the combined trade-offs between the utility, fairness, and explainability of models. These metrics are essential for quantifying and understanding the intricate balance required for responsible and unbiased algorithmic decision-making.

Finally, the majority of works focus on explanation methods aimed at understanding the causes of unfairness. There remains ample opportunity for the creation of further fairness metrics guided by these explanations and for leveraging them in the development of strategies to alleviate unfairness.

\section*{Acknowledgment}
This work has been partially supported by project MIS 5154714 of the National Recovery and Resilience Plan Greece 2.0 funded by the European Union under the NextGenerationEU Program.

\clearpage
\bibliographystyle{IEEEtran}
\bibliography{IEEEabrv,bibliography}

\end{document}